# Machine Learning Information Retrieval and Summarisation to Support Systematic Review on Outcomes Based Contracting

Iman Munire Bilal[1], Zheng Fang[1], Miguel Arana-Catania[2], Felix-Anselm van Lier[3], Juliana Outes Velarde[3], Harry Bregazzi[3], Eleanor Carter[3], Mara Airoldi[3], Rob Procter[1*]

*Affiliations:*
[1] *Department of Computer Science, University of Warwick and Alan Turing Institute for Data Science and AI, Coventry, United Kingdom*
[2] *Faculty of Engineering and Applied Sciences, Cranfield University, United Kingdom*
[3] *Government Outcomes Lab, Blavatnik School of Government, University of Oxford, Oxford, United Kingdom*
[*] *Corresponding author:* iman.bilal@warwick.ac.uk

Abstract

As academic literature proliferates, traditional review methods are increasingly challenged by the sheer volume and diversity of available research. This article presents a study that aims to address these challenges by enhancing the efficiency and scope of systematic reviews in the social sciences through advanced machine learning (ML) and natural language processing (NLP) tools. In particular, we focus on automating stages within the systematic reviewing process that are time-intensive and repetitive for human annotators and which lend themselves to immediate scalability through tools such as information retrieval and summarisation guided by expert advice. The article concludes with a summary of lessons learnt regarding the integrated approach towards systematic reviews and future directions for improvement, including explainability.

> Policy Significance Statement
> Systematic reviews, despite aiming to bolster evidence-based policy, often face challenges in linking research to practice due to their time-intensive nature and narrow scope. Using the example of a large-scale systematic review consisting of highly heterogeneous and complex social science literature, this research highlights the potential of machine learning (ML) and natural language processing (NLP) tools to speed up the efficacy of systematic reviews. In particular, we focus on automating stages within the systematic reviewing process that are time-intensive and repetitive for human annotators and which lend themselves to immediate scalability through tools such as information retrieval and summarisation guided by expert advice. In parallel, we highlight the need to evaluate the usability and understanding of automated tools by conducting a workshop and interviews with researchers and policy engagement professionals to understand areas of improvement for this integrated approach.

## 1. Introduction

Systematic reviews have become an important tool for researchers and policy-makers wishing to gain a comprehensive and rigorous understanding of a particular field. However, the volume of scientific papers and reports grows rapidly each year, making the process increasingly unmanageable for researchers and policymakers. The result is that researchers and policymakers often have to define searches very narrowly in order to limit the number of articles to that which is feasible for human analysis. Furthermore, systematic reviews become outdated very quickly. This significantly limits the "life" of systematic reviews and outdated reviews may mislead policymakers.

Because of these challenges, there has been increasing interest in the past 15 years in experimenting with using machine learning (ML) and natural language processing (NLP) techniques to support the systematic review process (Ananiadou et al., 2009). So far, however, research has mostly focused on screening abstracts and titles for full-text analysis (see e.g., van de Schoot et al. 2021). In addition, to date, studies that have applied ML techniques in systematic reviewing have largely been limited to the natural sciences (van Dinter et al. 2021).

In this project our principal aim has been to develop a proof-of-concept prototype to test how research can advance the application of ML and NLP techniques to meet the challenges posed by the growing volume of literature to the rigorous and timely execution of systematic reviews.

To address this, the project has focused on evaluating the application of state-of-the-art ML and NLP techniques in the context of an ongoing systematic review. This has been conducted by the Government Outcomes Lab (GO Lab) in partnership with Ecorys (Picker et al., 2021)[1]. Earlier work unveiled the potential of using computational tools for systematic reviews in the social sciences and the appetite of both policy makers and academics for such tools. In this project we have advanced and expanded our initial findings (Fang et al., 2024) by developing a basic prototype; tested and refined our methodology for subsequent research; and engaged researchers and policy-makers to provide feedback on the prototype.

In particular, the project set out to explore three key research objectives.

(1) **Test the viability of extending the application of machine learning tools to systematic reviews in the social sciences.** No research project has yet systematically attempted to translate the advances made in the use of ML for systematic reviewing to the social sciences. The social sciences present a unique challenge for machine learning-assisted systematic reviews. The literature in this field is diverse, encompassing various forms and content and addressing complex research questions with heterogeneous sets of research methods. As a result, traditional ML-assisted systematic review methods may not suffice, necessitating innovative approaches to data analysis.

(2) **Explore how ML can increase the efficiency and broaden the scope of systematic reviews in the social sciences.** Current machine learning tools in systematic reviewing focus mainly on the screening of titles and abstracts. They do not address the subsequent steps of analysing, comparing and synthesising full academic texts. This project pioneers data extraction at the full-text review stage of systematic reviews by surfacing data on emerging themes, trends, and patterns in the literature, extracting and summarising key elements of text, and identifying entities mentioned. The aim of this type of analysis is to (1) help researchers navigate large amounts of literature to conduct more efficient and targeted systematic reviews; to (2) create new datasets for novel methodologies and forms of analysis of systematic reviews; and (3) to create a visual interface making the information extracted from large corpora of academic and grey literature easily navigable for a broader audience, such as policy-makers.

(3) **Investigate requirements of ML-enabled tools that support the systematic reviewing process.** As machine learning has become more advanced, it has become more difficult for users to understand how these tools work, causing concerns about their trustworthiness and reliability. This may significantly hamper the value of ML. Research has identified promising techniques for explainable AI (xAI) to "open up" the black box, but it's unclear which techniques will be most effective for specific applications. As part of this, we evaluated the SyRoCCo tool (Fang et al., 2024), the prototype of which was developed by us prior to and continued in this project.

This project has followed a co-production approach, with the GO Lab team and ML & NLP group working together as a team throughout. Having users at the heart of the development process has long been recognised as the key to successful IT projects (Bodker et al., 1995). This has become especially important as machine learning techniques that are still so unfamiliar to many are now being widely applied to new products and services (Slota, 2020; Wolf, 2020). Objectives and progress were reviewed in weekly project team meetings. Initially, these focused on building common ground: (a) familiarising GO Lab team members with NLP and machine learning concepts, techniques, capabilities and limitations; (b) enabling the technical team members to gain an understanding of the review process; and (c) then establishing some initial requirements for how ML & NLP techniques and tools could support it. As the project progressed, the results of a series of experiments were reviewed, enabling the ML & NLP techniques and tools to be evaluated and progressively refined.

---

[1] The GO Lab is a research and policy centre based in the Blavatnik School of Government, University of Oxford, which investigates innovative ways in which governments partner with the private and social sectors to improve social outcomes; Ecorys is a economic research and consulting company.

The project has been divided in two main parts. First we conducted an evaluation of the existing machine learning tool for systematic reviews to understand its limitations, challenges and design its posterior developments. Second, based on the results of the previous evaluation we developed new machine learning functionalities, mainly focused on information retrieval, automatic summarisation, and explainability, and evaluated them.

## 2. Data and Case Study: A Systematic Review on Outcomes Based Contracting

This section outlines the systematic review process which we used to test ML and NLP tools to assist the reviewing process and make the review accessible to a broader audience.

The focus of GO Lab's and Ecory's systematic review is 'outcomes-based contracting' (OBC). OBC is a model for the provision of public services wherein a service provider receives payment, in-part or in-full, only upon the achievement of pre-agreed outcomes. There are multiple forms of OBC. It has been implemented in numerous countries, and applied to a range of policy areas, from education to environmental management (Government Outcomes Lab, n.d.). Approaches to OBC have been analysed and assessed both by researchers in an academic capacity as well as by evaluation consultants. As such, the empirical evidence on OBC is fragmented, dispersed, and difficult to navigate.

Additionally, OBC is an intrinsically interdisciplinary field of research, which contributes to the heterogeneity of the literature. OBC projects have been analysed and evaluated through a variety of disciplinary lenses, ranging from sociological to legal to economic. This is mirrored in the range of methodological approaches that have been used to assess OBC projects, and evidence encompasses methods from small-scale ethnographic analysis to large-scale quantitative economic analysis.

The purpose of GO Lab's and Ecory's systematic review is to gather and curate, for the first time, all of the existing evidence on OBC. Unlike conventional systematic reviews, the approach applied here is intentionally broad, including a variety of study designs and content. This broad-scale 'review of reviews' allows researchers to address multiple research questions with the same underlying set of studies. The ambitious size and scope of the review, as well as the heterogeneity of the literature, make it particularly suitable as a case to explore the benefits of applying machine learning.

The review's objective therefore entails dealing with a large and diverse body of published material. The initial search process in bibliographic databases returned over 11,000 results. After manual screening to establish their potential relevance to our topic, ~2,000 studies remained. To manually categorise and extract the details of a uniform set of variables from across this large and heterogeneous volume of publications is so labour-intensive as to be a practical impossibility. One way around this difficulty is to further filter the studies according to narrower inclusion criteria until the number of included texts becomes more manageable. Doing so leads to detailed enquiries into specific sub-themes, and this is indeed how the GO Lab and Ecorys managed the early policy reports produced from the systematic review evidence base (Bregazzi et al., 2022; Elsby et al., 2022). While this approach produces valuable insights for specific topics and policy areas, the manual filtering process is time consuming and leaves much potentially useful information inherent in the screened body of published material unaccounted for and inaccessible to a broader audience.

During the realisation of this systematic review, our team first developed a machine learning tool to assist in the review and present publicly the dataset called SyRoCCO[2]. This first version of the tool, prior to the current project, allowed to conduct several tasks on the OBC dataset: assign publications into 'policy area' categories; identify and extract key information for evidence mapping, such as organisations, laws, and geographical information; connect the evidence base to an existing dataset on the same topic; and finally identify subgroups of articles that may share thematic content. The tool also provided an interactive interface through which users could explore the data at ease, including several visualisations, such as a map, bar charts, and a similarity graph of related articles, as well as different filtering options.

---

[2] SyROCCo dataset https://zenodo.org/doi/10.5281/zenodo.12204303

As it was mentioned in the introduction, we conducted an evaluation of the tool during this project which led to the design of the new machine learning functionalities developed in the rest of the project. In the following section we will describe the initial evaluation of the tool.

## 3. SyRoCCO Tool evaluation

While the initial version of the SyRoCCO tool (Fang et al., 2024) has shown to yield substantial efficiencies for future systematic reviews and broaden their analytical scope, testing and assessing its usability through a deeper evaluation was necessary to improve the ease and confidence with which researchers and policy makers can access evidence. As we detail below, the evaluation surfaced key insights on necessary developments and refinements of machine learning techniques. These findings informed our strategy of improving the tool's capacity in terms of information retrieval, automatic scientific summarisation, and explainability.

This evaluation process involved the organisation of a workshop with potential users and a series of interviews with researchers and policy makers.

### 3.1. Evaluation workshop

As a first stage in the evaluation, we designed and delivered an evaluation workshop. The participants were 11 members of the Government Outcomes Lab team. Our team are experts in outcomes-based contracting, and include researchers and policy engagement professionals.

Participants were split into four groups, and each was given a vignette use case to prompt their exploration of the SyROCCo tool. The use cases sought to emulate questions or challenges that OBC practitioners and researchers might face in reality. The task for the groups was to use SyROCCo to answer those questions or solve those challenges[3]. A SyROCCo team member was assigned to each group to observe participants' use of the tool and answer questions. Their role was not to provide guidance or teach the group how to use the tool, however, as we wanted to observe how users interacted with the tool without being prompted or guided by someone with prior familiarity. The groups explored the tool for ~50 minutes. We then came together as a whole team, each group provided a 5 minute summary of their experience with the tool, and we discussed it together. In addition to collecting feedback from the participants, SyROCCo team members observed and took note of how participants interacted with the tool.

Based on their experience, participants made suggestions for how the design of the tool might be improved. These ranged from UI changes to additions that would require further development of the machine learning functions that underlie the tool and are the focus of this article. One of the key suggestions that oriented the rest of the project was generating a summary of the publications, such that the key points are extracted and presented to the user. More information about this was obtained during a posterior set of interviews, detailed in the next section.

Another relevant learning was related to the tool in terms of interaction and generalisability of the tool. Tools designed using NLP capabilities can provide many general functionalities to process and interact with the data. As it is the case here, it was found that the current design admits of wide variation in interaction - that is, there is not a narrowly defined 'route' through the tool's functions that users might generally be expected to follow. We therefore conclude that a user coming to this tool for the first time may not have an intuitive grasp of its functions, and may even remain unaware of some of its functions and capabilities. While NLP tools can provide very wide functionalities and be very generalisable, working on 'usability' and intuitive interactions is thus an important element of their design.

Some participants were confused by some aspects of the visualisation of the data. While NLP tools are traditionally focused on a textual interaction with the users, visual interfaces can help the explainability of the tool, but they have their own challenges, such as in this case to clarify which subset of the data is used in the visualisations, or creating visualisation around complex NLP complex metrics such as text similarity.

---

[3] The use cases included: a consultancy firm developing a report about the latest trends in outcomes-based contracting in India; a research team investigating networks of knowledge production around outcomes-based contracts; and a data activist writing a blog to comment on the complementarity between SyROCCo and data from the INDIGO Impact Bond Dataset.

Clarity is necessary to help people utilise all of the tool's functions to meet their needs, but it is also central to generating confidence in the tool's reliability. This was demonstrated by an incident in the workshop. One group used Google Scholar to search for a term that they had found in the tool's filter list ('disbursement-linked indicators'). On Google, they found some papers that were not in SyROCCo and they wondered why the tool had 'missed' these publications. This kind of observation might lead a user to suspect the reliability of the tool, if they were not aware of the systematic process of searching and screening that produced the tool's database. SyROCCo is not a search engine, it is a curated evidence base. But a user's first assumption may be that it is a search engine, and they may compare it unfavourably to other search engines (like Google Scholar, etc.).

It is therefore important that such tools include great clarity and explanations - not only of the functions themselves, but of the systematic review method and the machine learning process that underlie the tool. Confidence in SyROCCo - and in other tools that rely on machine learning - is crucial if it is to become a regular source of evidence among policymakers and researchers. But trust in its reliability might be undermined by a false understanding of what the tool is and does. Confidence can be pursued by providing clarity about what it is and what it is *not*. The challenge for such tools is to design the tool such that it is intuitive and user-friendly, whilst simultaneously making sure clarity and transparency is maximised such that users understand what they are working with.

### 3.2. Interviews

Following the previous workshop, we conducted six in-depth one-to-one interviews with practitioners, policy makers and researchers. The idea behind these interviews was to create a space where potential users of the tool who are not experts in outcomes-based contracting could share their views on the usability of SyROCCo. The pool of interviewees included a civil servant working to find innovative financing mechanisms for the government of Australia, a researcher investigating public sector contracting, a machine learning expert who previously worked on the creation of an outcomes contract dataset leveraging machine learning techniques and Google alerts, among other participants. All participants use data and evidence in their day-to-day job. There is great variety in the type of sources of evidence that they use. From Google searches to National Archives, from academic papers to banks of indicators (such as the World Bank Database), participants were well versed in the search of evidence to design better policies, to evaluate social programmes, to do research on the effectiveness of these programmes, among other tasks.  A detailed description of the participants' backgrounds can be found in Appendix D. The interviews were carried out as semi-structured interviews. We selected this type of interview as it was key to not only get feedback on SyROCCo, but also explore participants' thoughts and ideas around the use and potential of the tool.

Since the possible summarisation functionality was one of the main outputs of the previous workshop, participants were asked about their thoughts on the possibility of SyROCCo to offer a summary of a paper or report. Participants were enthusiastic about the idea and described what a good summary looked like. While researchers and users with an academic background focused on a similar formula of a good summary (research question, description of data, methodology and results), policy oriented users emphasised that a good summary should also highlight policy implications or recommendations coming from a paper.

All participants shared a list of suggestions for improvements. Some of the common recommendations included:

- Improve the explanation around key decisions behind the tool, such as how we allocate policy sectors to papers and what is our level of confidence on that decision.
- Improve the capacity of the tool to narrow down an initial search.
- Improve our description of the database behind SyROCCo and how it was built (thus, include more information on the systematic review process that took place before SyROCCo)
- Include more information that helps practitioners narrow down the search, so that practitioners can rapidly get a set of key papers to read, instead of a long list.

The evaluation highlighted the differences in informational needs of researchers and policy makers, which will need to be addressed in a future project.  The feedback suggests that policy makers require a more streamlined user experience with the tool, including easier navigation, faster loading times, and more specific information to narrow down search results. On the other hand, researchers need more academic parameters such as citation count and publication journal, as well as more descriptive columns and details about the methodology used in the papers.

### 3.3 Discussion on the evaluations

The feedback from the evaluation process suggested several strands of improvement to the tool. This included improved and more detailed information retrieval; summarisation of different elements of text; and better explanation systems that can improve the confidence of this tool. These challenges are particularly significant in the context of our case study due to the complexity and heterogeneity of the underlying data. Social science research covers a wide range of topics and methodologies, making it difficult to identify relevant papers and extract key information. Additionally, the language used in social science research can be highly nuanced and context-dependent, making it challenging for machines to accurately understand and summarise the text. Finally, creating better explanation systems for machine learning models requires a deep understanding of both the data and the underlying models, which can be difficult to achieve in a complex and rapidly evolving field like social science research.

### 4. ML and NLP techniques

In the second part of this project, we focused on testing different techniques to address the significant challenges of the use of machine learning tools for systematic reviews, motivated by the previous evaluation. The research was focused on information retrieval, automatic summarisation, and explainability. We describe the progress we made on these issues below.

### 4.1. Information Retrieval

Information Retrieval (IR) is the process of searching a large collection of documents for information that corresponds to a search query. In our project, IR is used to automatically locate relevant text passages related to our research questions. This includes questions relating to study design and target population, as well as issues more specific to outcomes based contracting, including financial aspects of an outcomes contract and the person-level outcomes achieved. The IR model receives two inputs: all text passages from our outcomes contract papers, and our research questions of interest. The model's output consists of a ranked list of text passages related to each specified research question. With these retrieved text passages, researchers can perform in-depth study on the question and text pairs.

In order to retrieve text passages relevant to the query, conventional IR strategies rely on hand-crafted features, where a feature is an input variable such as the frequency of a keyword (Robertson et al. 1994), the length of the query, or other salient indicators related to the task at hand. Features from both the query and text passages are used to compute the score of matching. One challenge is that these features tend to be domain specific and every time we shift to a new domain, a new set of specific features needs to be defined prior to model application. This is time consuming as new domain-specific features take time to validate.

To address this problem, the focus of IR research has shifted to neural retrieval approaches that accept raw text data as input and can automatically convert the text data into features suitable for a given IR task. Many different neural retrieval approaches (Karpukhin et al., 2020; Khatab and Zaharia, 2020; Yih et al., 2011; Lee et al., 2019; Karpukhin et al., 2020) have been proposed, including dense retrievers (Karpukhin et al., 2020), which can map both input queries and text passages to the same semantic space. The corresponding vectors in the semantic space are then used to compute the matching scores. Recent research has demonstrated, however, that these neural retrieval approaches struggle to perform effectively in a range of retrieval tasks without specialised training data (Thakur et al., 2021). Our project exemplifies this as we lacked appropriate training data. Microsoft introduced the MS MARCO dataset (Bajaj et al., 2016) to assist the IR research community in addressing this issue, by providing a dataset with a large scale and diversity to enable zero-shot transfer learning for a range of tasks. However, recent studies have shown that knowledge transfer from a single dataset alone is not guaranteed to bring benefits to all IR tasks and domains (Luiz Bonifacio et al., 2022). In our project, the documents of outcomes contracts have not been explored by IR tasks before and they are vastly different from those in the MS MARCO dataset, therefore the performance of general approaches trained on MS MARCO on our dataset is low.

Data augmentation is a technique used to increase the number of training samples. It has been widely used with small training sample sizes and has been successful in neural models with low rsource settings (Fadaee et al., 2017; Kobayashi, 2018). The recent developments of large scale pre-trained language models such as GPT-3 (Brown et al., 2020), BERT (Devlin et al., 2018) and FLAN (Wei et al., 2022) have further benefited data augmentation techniques and many works have shown that these large scale models can generate good quality synthetic data (Anaby-Tavor et al., 2020; Papanikolaou and Pierleoni, 2020; Yang et al., 2020; Mohapatra et al., 2021; Kumar et al., 2020; Schick and Schutze, 2021; Meng et al., 2022). Recently, Luiz et al., (2022) proposed a novel data augmentation framework which only requires a few simple prompts as the input to generate synthetic question and text-passage pairs. In this case, a prompt is formed by a question and its most relevant text-passage labelled by human annotators. This is suitable for our project since we only have access to very few manually annotated examples, i.e., question and text-passage pairs labelled by human experts in the GO Lab team. By augmenting the training dataset of our project, we can then train high-quality models to accurately find the text passages relevant to the research questions at hand.

### 4.2. Automatic scientific summarisation

Automatic summarisation has been extensively studied (Narayan et al.,2018; Liu & Lapata 2019; Lu et al. 2020) as a solution to understand huge volumes of data. With the rise of neural networks, models are able to generate summaries which contain salient information and reduce redundancies. It has been successful in fields such as news articles and product reviews which contain large-scale annotated datasets to satisfy the data-driven summarisation models. These domains also benefit from having accessible content which does not require experts for creation of the gold standard. In the past years, there has been consistent progress in adopting automatic summarisation of scientific literature by the introduction of new datasets (Yasunaga et al.,2019; Lu et al.,2020; Meng et al.,2021) and models (Chen et al.,2021; Soleimani et al.,2022). This falls into two categories: single document summarisation (SDS) or multi-document summarisation (MDS).

### 4.2.1. Single document summarisation

Given that academic articles contain an abstract which can already be considered a faithful summary of the article, single document summarisation (SDS) of academic works diverges in two directions: generate even shorter summaries or enrich the summary with more targeted information. For the first case, the task is inspired by extreme summarisation (Narayan et al., 2018) of news articles which can often be summarised by their first sentence. This has been adapted for the scientific literature by Cachola et al. (2020) who introduce the dataset SCITLDR and propose a baseline summariser that synthesises the abstract, introduction and conclusion of each paper in one sentence. Recent work by Mao et al. (2022) introduces CITES, a new benchmark for SCITLDR that is based on BART (Lewis et., 2020) and is trained on huge volumes of citation texts.

The other use case of SDS consists of generating longer summaries which are more informative than the abstract. Current work has done this by either producing a structured synthesis of the paper (Dong et al., 2021; Meng et al., 2021; Soleimani et al., 2022) or by enriching the current summary with the paper's future impact in the community (Yasunaga et al., 2019).

### 4.2.2. Multi-document summarisation

Multi-document summarisation (MDS) sets out to capture the information across documents by comparing and contrasting against similar topics. Automatically summarising groups of related documents is especially useful to researchers interested in writing surveys which are agreed to be very time-consuming (Liu et al., 2022) and thus would benefit from computational support. However, this type of summarisation is especially challenging for scientific articles due to two important reasons. First of all, the content comprises complex concepts, domain-specific acronyms, technical terms distinct to each domain. Secondly, the relations between papers are diverse and challenging to explain (Luu et al., 2021) which determine the way the information is introduced in the summary.

A popular way to formalise MDS is to generate summaries which take the form of the 'Related Work' chapter (Chen et al., 2021; Lu et al., 2020). Chen et al. (2021) construct an abstractive supervised summariser which makes use of the citation relationships between grouped papers. Unlike the previous approach, Lu et al. (2020) use a query document and its cited documents as the input for their summarisation model. Both proposed models produce summaries based on small (less than 5 documents per group) groups of documents from STEM disciplines which might impede the generalisability of the studies to other domains such as the social sciences.

### 4.3. Explainability of ML models

With the rise of neural networks and big datasets to support data-hungry models, much progress has been achieved in many NLP tasks: classification, automatic summarisation, information retrieval etc. However, most proposed models are 'black-boxes' which lack insight into their inner mechanism and impede the transparency and interpretability of their applications in downstream tasks. Providing explanations to a model's predictions can enhance users' trust and help with debugging in case the model prediction is incorrect (Ribera and Lapedriza, 2019).

Recently, research into Explainable AI has been prioritised with explanations covering many forms depending on the model to be explained and the task. Below, we present rationale-based and free-text explanations.

### 4.3.1. Rationale-based explanations

Rationales are the most prolific type of explanation with most works (Lu et al.,2022; Lei et al.,2016; Zhang et al.,2016) defining these as the subset of the input that contributes the most to the model prediction. The rationale is identified either intrinsically by the architecture of the model, usually relying on attention weights (Vaswani et al., 2017), or it is assigned by post-hoc scoring algorithms such as LIME (Ribeiro et al., 2016), Shapley values (Lundberg and Lee, 2017) or integrated layers (Sundararajan et al., 2017). A parallel body of work comprises counterfactuals (Tolkachev et al., 2022; Wu et al., 2021; Ross et al., 2021) which rely on rationales to construct minimal edits to the original input with the aim of generating a different prediction. Ross et al. (2021) argue that counterfactual explanations are more user-centered since human explanations benefit from contrastive examples for understanding.

### 4.3.2. Free-text explanations

Free-text explanations are written in natural language and are not constrained by instance inputs. Their definition is task-dependent which allows their application to many NLP fields with immediate applications to real-world use cases such as fact-checking (Alhindi et al., 2018), natural language inference (Camburu et al., 2018) and Question-Answering (Aggarwal et al., 2021). Besides improving the transparency of a model's decision, fluent explanations were also shown to improve model performance when used along-side the input for prediction tasks (Stammbach and Ash, 2020). This is argued by Teso and Kersting (2019) to discourage spurious correlations between input features learned by models. Despite their obvious appeal, the disadvantage of most mechanisms generating free-form explanations (Atanasova et al., 2020; Kotonya and Toni, 2020; Kumar and Talukdar, 2020) remains their reliance on gold justification authored by human annotators, the creation of which is a time-consuming process. There is current work (Bilal et al., 2024) focusing on generating fluent explanations based on summarisation in zero-shot settings or without supervision.

## 5. Methodology

### 5.1. Dataset

The GO Lab Team coded information from the papers in the corpus relevant to a set of predefined themes pertinent to the domain of outcomes-based contracting. Extracting these themes is a data and time-intensive stage of the systematic reviewing process and does not require specialised knowledge, making it a suitable task for automation via ML and NLP techniques. These themes fall into two main categories: state of evidence and synthesis of findings.

Four themes of the corpus were chosen for a series of experiments with ML and NLP techniques: Target Population (specific category), Study Design (specific category), Financial Details and Costs (broad category), Personal-level outcomes (broad category). Six papers were chosen for the experiments and were coded to provide the correct match for each of the 4 themes.

**Query selection**: The inputs of an IR model consists of text passages and queries of our interest. With the help of the GO Lab team, versions of the 4 research themes (Appendix A) were formulated to satisfy varying levels of fluency: stand-alone questions ('What is the study design?" etc.), keywords ("Study design"," methodology" etc.), and concatenation of keywords. These variations of the query are then validated to find the best performance setting for the IR model.

**Analysis of Retrieval Difficulty**: Articles were partitioned according to the level of difficulty faced by the retriever on a 3 point scale of *hard*, *medium* and *easy* (defined below). The analysis was carried out on the passages of each article for each query. This categorisation was very helpful as the patterning enabled researchers to understand the heterogeneity of papers that do not adopt standardised models for reporting interventions, such as PICO. The PICO tool focuses on the Population, Intervention, Comparison and Outcomes of a (usually quantitative article) (Methley et al., 2014). Although this approach is commonly used to describe studies in health (as it is commonly advocated across the field of evidence based medicine) across our corpus several of the key terms such as 'control group' and 'intervention' are either not relevant to qualitative research designs or are less-formally articulated.

We therefore sought to classify the level of retrieval difficulty based on the following principles:

- *Hard* paper if no tokenized paragraphs including the gold standard (GS) corresponding to that theme contain any of the curated keywords
- *Medium* paper if some tokenized paragraphs including the GS corresponding to that theme contain some of the curated keywords. In this case, we show the number of paragraphs which contain both the GS and the keywords out of the total number of paragraphs which contain the gold standard.
- *Easy* paper if most of the tokenized paragraphs including the GS corresponding to that theme contain some of the curated keywords. In this case, we show the number of paragraphs which contain both the GS and the keywords / the total number of paragraphs which contain GS.

We find that classifying the articles in our sample proves especially helpful for hard and medium papers as this provides clarification in determining if the difficulties were caused by inappropriate keywords or tokenization techniques used in the precursor stages of the process. In addition to assigning difficulty labels, we also include a detailed explanation of each paper, which aids us in identifying the limitations of our ML tools and paves the way for the right development direction.

## 5.2. Information retrieval

In this section, we present our experiments in information retrieval. Recall that IR is used to locate text passages that are relevant to our research questions. By introducing methods to automatically locate relevant text passages, it can reduce the time cost of researchers reading papers and increase their efficiency in conducting systematic reviews.

We follow a two-step IR pipeline. In the first step, a retrieval model is used to retrieve a number of candidate passages relevant to the input query and ranked by their relevance. In the second step, these candidate passages are re-ranked by a re-ranking model to provide a more accurate ranking.

For the first experiment, we tested different models trained on large scale diverse datasets to verify if the general approaches trained on external datasets (as the MS MARCO case mentioned before) are applicable to our dataset. We observe that the documents of outcomes contracts have not been explored through IR tasks before – the format (grey literature vs articles following a traditional fixed structure) and content (specialised knowledge vs general knowledge) of these is considerably different from the documents in the MS MARCO dataset, thus posing an additional challenge in our experiments.

For the second experiment, we implemented the data augmentation framework from Luiz et al. (2022) to test if synthetic data could help improve the performance of IR models on our dataset. The framework is suitable for our project as it only requires a few simple prompts as the input. Here, a prompt is formed by a question and its most relevant text-passage labelled by human annotators. In our case we only have very few supervised examples, consisting of question and text-passage pairs labelled by human experts in the GO Lab team. Both experiments aim to retrieve the most relevant passages in the outcomes contracts for each research query.

### 5.2.1. Experiment on IR models trained on external datasets

**Setup**: In this setup, we tested models trained on large scale diverse datasets to check if the general approaches trained on external datasets are applicable to our dataset. We considered two types of text-level granularity for the passages retrieved by the model: paragraphs and sentences. For paragraphs, we tokenized each paper on outcome contracts to paragraphs and used these paragraphs as the text passages. For sentences, we further tokenized each paragraph to sentences as the text passages.

**Models**: We implemented two different model settings.
- Model 1: using 'multi-qa-mpnet-base-dot-v1'[4] as the retrieval model and 'cross-encoder/ms-marco-MiniLM-L-6-v2'[5] as the reranking model. Both are built upon the sentence transformers architecture. 'multi-qa-mpnet-base-dot-v1' was pre-trained on 215M (question, answer) pairs from diverse sources and has the best IR performance over 6 datasets compared with other models[6]. 'cross-encoder/ms-marco-MiniLM-L-6-v2' was pre-trained for the reranking purpose, and it has the second best performance on the TREC Deep Learning 2019[7] IR task and the MS Marco Passage Reranking LeaderBoard[8] (only 0.01 percent lower than the large version of it and thus more efficient in light applications) compared with other models[9].
- Model 2: Using the CoT-MAE retrieval model[10] and the reranking model[11] provided by Wu et al. (2022), which achieved a high ranking in the MS Marco Passage Reranking LeaderBoard, and the code is accessible to the research community.

---

[4] https://www.sbert.net/docs/pretrained_models.html#sentence-embedding-models
[5] https://www.sbert.net/docs/pretrained-models/ce-msmarco.html
[6] https://www.sbert.net/docs/pretrained-models/ce-msmarco.html

**Evaluation**: We used precision@k and recall@k to evaluate the performance of different IR models. Precision@k measures the percentage of correct matches among the top k retrieved passages. Recall@k measures the percentage of correct matches retrieved versus all correct matches in the entire dataset. Higher scores indicate better retrieval ability. We vary the value of parameter k from 1 to 20. We observed that the difference between two model settings is not significant when the value of k is small, but when we increased k to 10 and 20, the Model 1 setting achieved better recall scores, which means it can retrieve more relevant text passages from the entire dataset. However, there are still many relevant passages missed by the model, as we find that some papers return no relevant passages (suggested by the recall scores = 0). We also observed that the recall scores for different papers are consistent with our analysis of the difficulty of retrieval for different papers, as we are able to obtain higher recall scores for easy papers, while we usually obtain lower recall scores for difficult papers. We also report the results of replacing paragraphs with sentences as text passages, but we didn't observe significant improvements.

### 5.2.2. Experiment on IR models trained on augmented datasets

**Setup**: In this setup, we tested the data augmentation framework inPars[12] (Luiz et al., 2022) to measure if the augmented dataset could help improve the performance of IR models. The training set generated by inPars comprises triples consisting of: a question, a text passage containing the answer to the question (denoted as positive), and a text passage not containing the answer (denoted as negative). We explain the process in detail below.

The generating process of inPars involves four steps. In the first step, a question generating model is used. It takes simple prompts formed by a few (question, text passage) training pairs to generate new questions for other text passages in the dataset. In the second step, it filters the resulting synthetic (question, passage) pairs by a scoring function to obtain high-quality pairs. These high-quality pairs are then used as the positive examples. In the third step, it follows Pradeep et al. (2021) to create a training set from these positive examples by generating negative passages for each question to form positive-negative triplets. In the final step, we train a MonoT5 (Raffel et al., 2020, Nogueira et al., 2020) reranking model represented by a binary classifier that uses the positive-negative triplets as the training signal. We used questions and their corresponding text passages labelled by the GO Lab team to create prompts (Appendix B), and employed BM25 (Robertson et al., 1994) as the retrieval model to obtain candidate text passages and the trained MonoT5 neural reranking model to rerank these candidates. Metrics for precision@k and recall@k are used to evaluate the performance of the proposed framework on our dataset.

**Models**: There are different models that can use prompts to generate questions for input text passages. We benchmarked two popular models: OpenAI GPT-3 and Google FLAN.
- GPT-3 is a well-known generative language model, which at the moment of conducting this research was the most advanced version of the OpenAI GPT models. The advantage of using GPT-3 is that it can be accessed through the OpenAI search API without any need for local loading or powerful GPU resources. machines. The disadvantage to this approach is that the API usage is expensive when applied on large-scale corpora. We present the price of different GPT-3 models when this study was conducted in Table 1. For our purpose, we chose GPT-3 Curie, which is the second most capable model and costs $0.006/1K tokens (tokens are pieces of words where 1,000 tokens≈750 words).

---

[7] https://microsoft.github.io/msmarco/TREC-Deep-Learning-2019

[8] https://github.com/microsoft/MSMARCO-Passage-Ranking/

[9] https://www.sbert.net/docs/pretrained-models/ce-msmarco.html

[10] https://huggingface.co/caskcsg/cotmae_base_msmarco_retriever

[11] https://huggingface.co/caskcsg/cotmae_base_msmarco_reranker

[12] https://github.com/zetaalphavector/inpars

| Model | Training | Usage |
| --- | --- | --- |
| Ada | $0.0004/1k tokens | $0.0016/1k tokens |
| Babbage | $0.0006/1k tokens | $0.0024/1k tokens |
| Curie | $0.0030/1k tokens | $0.0120/1k tokens |
| Davinci | $0.0300/1k tokens | $0.1200/1k tokens |

**Table 1. Usage costs of different OpenAI Language models.**

- FLAN is also a generative language model. A FLAN model with 137B parameters has been reported to outperform a GPT-3 model with 175B parameters on various NLP tasks (Wei et al., 2022). The major advantage of using FLAN is that it is freely accessible. However, the drawback to this is that due to hardware limitations, we can only use the 3B parameters FLAN model, thus sacrificing model parameter size.

For both models, we set the maximum number of characters an input document must have to 4000, and the sampling temperature to 0. We used the default parameters setting from the original code for other parameters. When using Curie to generate questions, we set the training epoch of the MonoT5 neural reranking model to 1. When using FLAN for generation, we set the training epoch to 10, because we found that the T5 model failed to converge with only 1 training epoch.

**Results**: We present the full results in Appendix C. Model 3 represents using inPars with GPT-3 Curie as the question generating model, and Model 4 represents using inPars with FLAN as the question generating model. For Model 3, due to the expensive cost of Curie, we only generated questions for 6,791 paragraphs from our outcomes contract documents. After filtering for high quality question-paragraph pairs, we have 164 samples left. We then generated positive-negative triplets based on the 164 samples and trained a MonoT5 neural reranking model over the triplets. We used BM25 as the retrieval model to retrieve candidate text passages and the trained MonoT5 reranking model to rerank these candidates. For Model 4, we are able to generate questions for all 324,003 paragraphs from our documents, as the model is free to access. We have 196 samples left after filtering high quality question-paragraph pairs and generated positive-negative triplets based on the samples.

We computed precision@k and recall@k based on the reranked text passages. Compared to the setup of the previous two models with no synthetic data in the first experiment, we observed considerable improvements in Model 3 and Model 4, with five of the six papers achieving perfect recall scores (1.00) for the "study design" research theme, as well as varying degrees of improvements for the other research themes. As such, it can be concluded that for our dataset, data augmentation approaches are more effective than general approaches trained on external datasets. We have also not observed significant differences between Model 3 and Model 4, which can be explained by the similar order of magnitude in their parameter size: the GPT-3 Curie model contains 6.7B parameters while the FLAN model we used contains 3B parameters. For future work, hardware resources with considerable GPU memory are necessary to enable the testing of larger models.

**Explaining divergent model performance:** It became apparent during the development and testing of the IR models that recall performance varied by publication. This led the ML team to designate publications as *easy*, *medium* or *hard*, depending on how well the IR models were able to identify the most relevant text passage (gold standard) for each of the four test themes. GO Lab researchers reviewed the results from this categorisation of difficulty, and sought to ascribe a preliminary explanation for each. We considered the reason *why* a particular publication was *hard* for the model to identify the target population, for example, or why another was *easy* for the model to identify the study design, and so on.

While the small sample size does not allow us to present our explanation as patterns or trends, it has nevertheless raised some hypotheses for explaining model performance that can be investigated further and tested more rigorously in the next period of the project. These hypotheses are as follows:

- *Terminology.* As expected, the extent to which the terminology of the queries matches that of the publication may be one explanation for divergence in model performance.
- *Formatting.* The formatting of the document appears to influence information retrieval. Content presented in tables, for example, seems more difficult to extract than plain text. And documents which have been heavily formatted with graphic design similarly present difficulties.
- *Complexity of the relevant content.* Concise and relatively simple content may be easier for the IR model to identify than that which is detailed and complex. For example, a research design with a single aspect could be easier to identify than a more complex research design that features multiple methods and requires lengthier explanation.
- *Length of the document.* Even if the relevant content is clearly and unambiguously stated within a text, the model may struggle to pick out the precise sentence(s) required when a document is hundreds of pages long. This could be characterised as a problem of 'finding a needle in a haystack'.
- *Propensity for misleading content.* Finally, content on a particular theme may be easier for the model to find if the associated terminology is distinct - meaning the model may be less likely to be misled into picking up irrelevant content that is similarly phrased or uses some synonymous terms.

The evidence base for our systematic review is heterogenous, consisting of documents of different lengths, formats, and methods, and with low standardisation of reporting across the corpus. As such, it is important to gain a greater understanding of which kinds of publication are more or less amenable to information retrieval by machine learning tools. It has implications for the future reliability of machine learning supported systematic reviews. Formalising the assessment of difficulty may help indicate instances when we can be confident in a model's results, and where the results may benefit from human validation.

### 5.2.3. Discussion

The purpose of the information retrieval experiments described above was to identify appropriate techniques to automatically locate text passages related to the research themes of interest in each paper. This could help speed up the systematic review process. By automatically providing highly relevant text passages to the research themes, researchers do not need to read the full paper to manually identify relevant text snippets about the themes. This would save time and enable effective in-depth analysis to understand integrated questions and text pairs. In addition, this would also promote our automatic scientific summarisation work introduced in the later section.

Two different types of techniques were tested, techniques trained on large external datasets specifically designed for information retrieval tasks, and techniques trained on datasets augmented from our own dataset. The evaluation results confirm that the general approach of training on external datasets is not suitable for our outcomes contracts dataset, since most of the documents in the dataset are not publicly accessible. In other words, the data-enhanced approach is more feasible when developing information retrieval techniques for rare texts.

The close collaboration between the ML team and the GoLab team during the experiments also ensured the project moved in the right direction. During model selection and validation, the ML team performed multiple rounds of manual checks. The retrieval difficulty analysis performed by the GoLab team was also important for understanding the retrieval performance between different papers. This analysis offers potential avenues of how to employ information retrieval tools effectively to support the systematic review process whilst ensuring reliability of the results. For example, we envisage a human-in-the-loop system in which the tool recognises potential low accuracy of information retrieval and flags articles that require human input for reliable data extraction.

### 5.3. Automatic scientific summarisation

The summarisation experiments were based on the use case of supporting in-depth analysis of the corpus by providing document-level summaries of each research theme. The gold standard summary is provided by the GO Lab team and serves as a benchmark in our human evaluations of automatic summarisation models.

In the first experiment, we tested the ability of summarisation models to synthesise information relevant to a particular theme based on what was available in the paper abstracts. This was followed by a second experiment with an improved set of automatic summarisers that summarise targeted information obtained from the full text body of documents.

### 5.3.1. Experiment on summarisation of abstracts

**Setup**: Automatic summarisation was initially attempted for the 'Study Design' theme on a test set of 5 articles. The summaries were generated on an article-level and leveraged information from the abstract.

**Baselines**: Because the gold standard for Study design is defined by the human experts in GoLab as a 'one-sentence description', we aimed to choose models which were either designed for extreme summarisation (one-sentence summary) or have shown good performance for zero-shot settings. The chosen summarisation baselines are: T5[13] (Raffel et al., 2020), an abstractive transformer-based model; BertSumExtAbs[14] (Liu & Lapata, 2019), a two-stage extractive-abstractive model trained on the XSum dataset for extreme summarisation; and CATTS[15] (Cachola et al., 2020), a pre-trained model on scientific articles which employs titles as auxiliary training signals.

**Results**: When compared to the gold standard for this research theme, all candidate summaries were found unsuitable by the evaluators as they do not provide information relevant to the theme. Upon further investigation, this result is explained by the fact that the abstract suffers from the same limitation, thus causing any summary of the abstract to be insufficient. Additionally, we note that the version of BertSumExtAbs pre-trained on XSum (Narayan et al., 2018) mostly generates summaries which contain hallucinations (false information introduced by the model). This is in line with the finding by Cachola et al. (2020) who advise that the concept of extremely short (also referred to as TLDR) summaries can vary across domains: in our case, extreme summarisation for news articles does not fit the informational needs required by summaries of outcomes-based contracting articles.

This experiment motivated our progress to the information retrieval stage, which aims to extract only passages relevant to a research theme from the articles. After information retrieval was conducted, summarisation was then attempted again using an improved text input.

### 5.3.2. Experiment on summarisation of relevant passages

**Setup**: The second experiment had the same aim of producing summaries for the 'Study Design' theme. We note that several aspects about the methodology have changed. The test set of documents was expanded to 10 to improve the robustness of our results. The task now benefited from an improved input for summarisation as this factor contributed the most to the low performance of models in Experiment 1. Instead of relying on abstracts alone, we now have access to the subsets of text passages (highlights) from each paper which were used for the creation of the gold standard summary. These were provided by the GO Lab team. Note that the automation of this step is non-trivial as the relevant highlights are spread out across the articles and employ rich, sometimes technical vocabulary; this task is carried out by the IR stage independently.

**Baselines**: The choice of baselines for this experiment was informed by recent progress in the field of summarisation. We selected CITES[16] (Mao et al., 2022), the current state-of-the-art model for extreme scientific summarisation (outperforming the previous CATTS (Cachola et al., 2020) model for academic inputs), GPT-3 (Open AI) language model which has shown promise for its question generation performance in the information retrieval step and finally, BART[17] (Lewis et al., 2020) fine-tuned on the XSum dataset, which has known success in many summarisation tasks. In this experiment, we ran two iterations of evaluations to filter the best candidate summary possible for our task: in the first evaluation 2.1, we compared off-the-shelf versions of CITES (both its TLDR version as well as its version for title generation), BART and GPT-3 to rank the best two models. For evaluation 2.1, we further fine-tuned and adjusted the hyper parameters to ensure that the performance of our final models was maximised. In particular, we follow the advice of CITES' authors that have demonstrated an impressive performance boost through few-shot learning for new-domains. Consequently, we fine-tuned CITES on a set of 31 papers (disjoint from the test set) covering both academic and grey literature.

---

[13] https://huggingface.co/t5-base
[14] https://github.com/nlpyang/PreSumm
[15] https://github.com/allenai/scitldr
[16] https://github.com/morningmoni/CiteSum
[17] https://huggingface.co/facebook/bart-large-xsum

**Evaluation**: The evaluation was carried out in a systematic manner. Inspired by the Best-Worst technique (Orme., 2009), the assessment was set up by first constructing tuples of summaries about the same article from each model. We asked the evaluators to identify the best and second best model in each tuple and we score the models with 1 and 0.5 respectively each time they are chosen. Following this procedure, we produced a score (the sum across tuples) for each model. When shown to the evaluators, the names of the models responsible for each candidate were not disclosed. To avoid any potential positional bias, we randomly shuffled the positions of model candidates within the tuples (i.e. summary generated by Model A should not appear in a fixed position across tuples).

**Results**: The results of evaluation 2.1 are shown in Table 2. The agreement between the 3 annotators is calculated as the average Krippendorf's alpha = 0.79 which is interpreted as marginally very good agreement. The best two baselines are by far GPT-3 and the CITES model trained for extreme summarisation.

| **CITES (TLDR)** | **BART (XSUM)** | **CITES (Title)** | **GPT-3** |
|---|---|---|---|
| 5.50 | 0.83 | 0.83 | 6.83 |

Table 2: Results of evaluation 2.1

The performance of CITES (TLDR) is explained by the compatibility between its train and test domains; the model has been trained to generate succinct summaries for scientific STEM papers and it is tested on academic domains. Though there exists a slight domain shift between STEM and Social Sciences, Mao et al. (2022) note that the CITES model performs on par with supervised baselines in zero-shot settings.

We additionally received feedback from the evaluators that summaries were prematurely 'cut short', which is a sign that we need to increase the maximum token limit of the generation step; this is an aspect we amended in the next iteration.

The results of evaluation 2.2 are shown in Table 3. The agreement score is now average Krippendorf's alpha=0.67, lower than in the previous evaluation. This lower inter-agreement score is intuitive since the task was now much more nuanced due to the model filtering step of low-performing models. In fact, feedback received from the annotators confirmed the increased difficulty of this second round, in particular in distinguishing between the best and second best summary.

| **CITES** | **CITES (fine-tuned)** | **GPT-3** |
|---|---|---|
| 2.66 | 6.5 | 6.5 |

Table 3: Results of evaluation 2.2

For this round, we experimented with CITES fine-tuned on a sample of 31 documents and their corresponding gold summaries, and off-the-shelf CITES and GPT-3 for which we extended the allowed summary length from 50 tokens to 60 tokens.

We found that the fine-tuned CITES and GPT-3 models performed equally well. This is similar to the results of Goyal et al. (2022) who observe that zero-shot GPT-3 is a competitive baseline even compared against fine-tuned summarisers. We note that CITES' performance can be potentially improved by extending the fine-tuning sample to the proposed size (128 manually-written summaries in test set) advised by the authors. However, there is a trade-off between performance and supporting the system review process through our methods. Our aim is not to achieve the best scores (possible by training our models on the entire dataset and involving extreme annotation effort), but to assist researchers and policy makers in understanding their corpus.

Another important remark is that although GPT-3 obtains high performance without supervision, the model involves running costs. We employed the 'davinci' version prevalent in recent work (Goyal et al., 2022; Bhaskar et al., 2022) for the summarisation task, which is also the most expensive option offered at that moment by OpenAI.

### 5.3.3. Discussion

Our aim in the aforementioned experiments was to assist the systematic review process during its initial synthesis of individual papers across significant research themes which had been identified prior to our project. In practice, this is done by automatically generating a targeted document-level summary that informs a user about the document's contents with respect to a theme.

Our results on automatic summarisation confirmed the importance of close collaboration between the ML and GoLab teams to produce a fine-tuned summarisation baseline. This is done by training our summariser on a small sample of human-authored summaries which help the model 'learn' the important patterns to synthesise for the specific theme. To reach our conclusion, human evaluations of this fine-tuned model and other competitive baselines are conducted with model performance and running costs considered.

Note that, although we focused on a specific research theme ('Study Design') and on a sample of papers due to time constraints, we remark that our methodology can be reproduced for the entire corpus and can be tailored for each research theme.

## 6. Conclusions and future work

In this section we summarise what we have learnt from the experiments in information extraction and summarisation in terms of future work directions.

### 6.1. Information Retrieval

Future work will first focus on tuning the prompts to augment the training dataset. As we introduced before, the generating process of inPars involves taking prompts to generate new questions for other text passages in the training dataset. Prompts served as the training signal and high quality prompts would help direct inPars to generate high quality synthetic data and improve IR performance. This will be achieved by testing different question-passage pairs to create prompts and find questions that result in the best retrieval performance.

Another direction is to associate the extracted information with the different outcomes contracts described in the document. A document usually contains analysis of multiple outcomes contracts. Currently, we have not taken into account these different outcomes contracts and do not have labelled information to indicate which retrieved text passages are associated with which outcomes contract.

Another potential direction is to utilise inPars to automatically generate reasonable new questions for the research themes of interest. Currently, we are using manually created questions to retrieve information, which may be biased due to knowledge limitations. If we can provide machine-generated questions that reveal the corpus from different perspectives, then researchers can draw inspiration to analyse the documents from more lenses other than the established ones.

### 6.2. Summarisation

Future extensions of the work will first concentrate on tailoring our summarisation strategy to cover all themes relevant to our corpus (e.g. 'Target Population', 'Financial details', etc.). Our current approach can be easily adapted for these new themes by fine-tuning the model on small samples of gold standard summaries specific to the information needs of each category. These targeted summaries can also be expanded to cover multiple documents and contrast findings across multiple themes. However, this extended scope would require a close collaboration with the GoLab team to define the roles a multi-document summary should fulfill.

A pivotal future direction will be constructing a pipeline connecting the information retrieval stage to the summarisation stage. Up to this point, the models have exclusively used the highlights selected by the GoLab team from the source documents. However, this is a time-consuming task as each category requires the detection of its own specific set of highlights in each paper. The goal is to integrate the retrieved passages resulting from the extraction component into the input of the summarisation component without sacrificing the relevancy of the final summaries.

### 6.3. Explainability

Future work will focus on providing explanations to the ML tools we have employed, in particular, to the information retrieval component. Understanding the behaviour of the extraction model is prioritised because its outputs will inform further stages and initial errors in its predictions might propagate across the proposed work pipeline.

The first challenge is the modularity of the best performing models for the information retrieval task (Model 3 & Model 4) which involves synthetic data generation, retrieving documents and then reranking them via a fine-tuned model. Ideally all these steps should be validated by explainability experiments which makes the task time-consuming.

The second challenge is adapting the IR model to existing explainability techniques. Due to a paucity of explainable IR research for scientific documents, we propose using classification as a proxy for the extraction task: top-k documents labelled as relevant (for a predetermined k variable) and all others labelled as irrelevant. This is in line with work by Singh et al. (2019) that rely on label probability distributions constructed for the pseudo-classifiers according to ranks or retrieval scores of documents. This choice would enable us to use explainability techniques (LIME, Integrated Gradients etc.) which are widely available for classification tasks.

We are also interested in pursuing the full potential IR-based explainability by contributing an inherently explainable IR baseline to our task. Making use of the full aspects of IR (rank of documents, retrieval scores) is the ideal way forward to understand model performance. For example, using counterfactuals to explain why a document was ranked in its position with respect to higher scored results is a question we would like to answer in future experiments.

**Acknowledgments.** Rob Procter would like to acknowledge the support of the Alan Turing Institute for Data Science and AI through his Turing Fellowship, and the Government Outcomes Lab, Blavatnik School of Government, Oxford University, through his visiting Professorship.

**Funding statement.** We thank these funders for their contribution to this work: Department for Digital, Culture Media & Sports (grant ref: A2683); Children's Investment Fund Foundation (grant ref: 2104-06351); UK Research and Innovation (grant ref: MR/T040890/1); the UK Foreign, Commonwealth and Development Office (grant ref: 300539); UBS Optimus Foundation (grant ref: 51962); and the John Fell Fund (grant ref: 0012257).

**Competing Interests.** None.

**Data availability statement.** The code is available in the Github repository https://github.com/bilaliman/systematic_reviews. The data is available to preview at https://golab.bsg.ox.ac.uk/knowledge-bank/indigo/syrocco-ml-tool where more information about individual papers within the studied dataset can be found.

**Author contributions.**
Iman Munire Bilal: Conceptualization, Methodology, Software, Validation, Investigation, Writing - Original Draft, Visualization.
Zheng Fang: Conceptualization, Methodology, Software, Validation, Investigation, Writing - Original Draft, Visualization.
Miguel Arana-Catania: Conceptualization, Methodology, Software, Validation, Investigation, Writing - Original Draft, Visualization, Supervision, Funding acquisition.
Felix-Anselm van Lier: Conceptualization, Methodology, Software, Validation, Investigation, Resources, Data Curation, Writing - Original Draft, Visualization, Supervision, Funding acquisition.
Juliana Outes Velarde: Conceptualization, Methodology, Software, Validation, Investigation, Resources, Data Curation, Writing - Original Draft, Visualization, Funding acquisition.
Harry Bregazzi: Conceptualization, Methodology, Software, Validation, Investigation, Resources, Data Curation, Writing - Original Draft, Visualization, Funding acquisition.
Eleanor Carter: Conceptualization, Methodology, Software, Validation, Investigation, Resources, Data Curation, Writing - Original Draft, Visualization, Supervision, Funding acquisition.
Mara Airoldi: Conceptualization, Methodology, Software, Validation, Investigation, Resources, Data Curation, Writing - Original Draft, Visualization, Supervision, Project administration, Funding acquisition.
Rob Procter: Conceptualization, Methodology, Software, Validation, Investigation, Writing - Original Draft,

Visualization, Supervision, Project administration, Funding acquisition.

Dong, Y., Mircea, A., & Cheung, J. C. (2020). Discourse-aware unsupervised summarization of long scientific documents. arXiv preprint arXiv:2005.00513.

Fadaee, Marzieh, Arianna Bisazza, and Christof Monz. "Data augmentation for low-resource neural machine translation." arXiv preprint arXiv:1705.00440 (2017).

Fang, Z., Arana-Catania, M., van Lier, F. A., Velarde, J. O., Bregazzi, H., Airoldi, M., Carter, E., & Procter, R. (2024). SyROCCo: Enhancing Systematic Reviews using Machine Learning. arXiv preprint arXiv:2406.16527.

Goodwin, T. R., Savery, M. E., & Demner-Fushman, D. (2020, December). Flight of the pegasus? comparing transformers on few-shot and zero-shot multi-document abstractive summarization. In Proceedings of COLING. International Conference on Computational Linguistics (Vol. 2020, p. 5640). NIH Public Access.

Goyal, T., Li, J. J., & Durrett, G. (2022). News summarization and evaluation in the era of gpt-3. arXiv preprint arXiv:2209.12356.

Karpukhin, V., Oğuz, B., Min, S., Lewis, P., Wu, L., Edunov, S., Chen, D. and Yih, W.T., (2020, November). Dense Passage Retrieval for Open-Domain Question Answering. In Proceedings of the 2020 Conference on Empirical Methods in Natural Language Processing (EMNLP) (pp. 6769-6781).

Khattab, Omar, and Matei Zaharia. "Colbert: Efficient and effective passage search via contextualized late interaction over bert." Proceedings of the 43rd International ACM SIGIR conference on research and development in Information Retrieval. 2020.

Kobayashi, Sosuke. "Contextual augmentation: Data augmentation by words with paradigmatic relations." arXiv preprint arXiv:1805.06201 (2018).

Kotonya, N., & Toni, F. (2020, November). Explainable Automated Fact-Checking for Public Health Claims. In Proceedings of the 2020 Conference on Empirical Methods in Natural Language Processing (EMNLP) (pp. 7740-7754).

Kumar, Sawan and Talukdar, Partha. 2020. {NILE} : Natural Language Inference with Faithful Natural Language Explanations. In Proceedings of the 58th Annual Meeting of the Association for Computational Linguistics

Kumar, Varun, Ashutosh Choudhary, and Eunah Cho. "Data augmentation using pre-trained transformer models." arXiv preprint arXiv:2003.02245 (2020).

Lee, Kenton, Ming-Wei Chang, and Kristina Toutanova. "Latent retrieval for weakly supervised open domain question answering." arXiv preprint arXiv:1906.00300 (2019).

Lewis, M., Liu, Y., Goyal, N., Ghazvininejad, M., Mohamed, A., Levy, O., Stoyanov, V. and Zettlemoyer, L., (2020, July). BART: Denoising Sequence-to-Sequence Pre-training for Natural Language Generation, Translation, and Comprehension. In Proceedings of the 58th Annual Meeting of the Association for Computational Linguistics (pp. 7871-7880).

Liu, S., Cao, J., Yang, R., & Wen, Z. (2023). Generating a structured summary of numerous academic papers: Dataset and method. arXiv preprint arXiv:2302.04580.

Lu, Y., Dong, Y., & Charlin, L. (2020). Multi-XScience: A large-scale dataset for extreme multi-document summarization of scientific articles. arXiv preprint arXiv:2010.14235.

Lundberg, S. M., & Lee, S. I. (2017). A unified approach to interpreting model predictions. Advances in neural information processing systems, 30.

Luu, K., Wu, X., Koncel-Kedziorski, R., Lo, K., Cachola, I., & Smith, N. A. (2021, August). Explaining Relationships Between Scientific Documents. In Proceedings of the 59th Annual Meeting of the Association for18

**Appendix A: IR queries**

**Study design**

A fairly specific category, by 'study design' we aim to capture the method used by the researchers for each paper/report

Keywords:

    Study design; method; methodology; data collection; research design

Key questions:

    What is the study design?; what is the research method?; How was data collected and analysed?

**Target population**

Another specific category, 'target population' refers to the numbers and characteristics of the people for whom the SOC aims to achieve outcomes

Keywords:

    Target population; beneficiaries; service users; participants; eligible population; eligibility criteria; cohort; clients

Key questions:

    What is the target population?; who are the intended beneficiaries of the service?; who does the service try to help?; Who was eligible for inclusion in the intervention?

**Financial detail and costs**

This is a broader category that could encompass a few different aspects. For our purposes we are likely most interested in the contract value, the amount paid for outcomes achievement, and the extra costs or savings achieved by using an OBC. As there are a lot of impact bonds in the education reports, I wouldn't be surprised if *investment* detail also gets captured by the machine learning tools. A further note - sometimes *time* costs are also reported, so that might also get pulled out by the machine here.

Keywords:

    Outcomes payment; price; contract value; contract cap; rate card; incentive payment; costs; savings; outcome pricing

Key questions:

    What are the costs of the contract?; how much is paid for outcomes?; what are the outcomes payments?; what is the total contract value?; what is the price per outcome?

**Person=level outcomes of the SOC**

This is also a broad category, and falls under our interest in 'SOC-effect'. Each SOC specifies specific outcomes metrics that the project is trying to achieve. 'Person-level outcomes' refers to any reported results about the achievement/non-achievement of these outcomes metrics. The reporting of outcomes can be quite diverse, with the amount of detail provided varying. You may get a detailed breakdown of quantitative quasi-experimental results, or it could just be a simple couple of sentences. This may prove tricky for the machines, but let's see!

Keywords:

>Results; outcomes achieved; impact; achievements

Key questions:

>What outcomes were achieved?; what impact was achieved; what were the results of the intervention?; what was the impact of the intervention?; were the contracted outcomes achieved?

**Appendix B: IR prompts**

Example 1:
Document: This study used a sequential explanatory equal-status mixed-method design to investigate whether Maryland's child care tiered reimbursement system incentivized child care centers to be rated at least 3 (and receive an incentive payment) on Maryland's 5-level Quality Rating and Improvement System (QRIS). The first stage of research consisted of multilevel logistic regressions to determine the association between centers' reliance on child care subsidy payments and whether the center had a rating of 3 or higher.
Relevant Query: What is the research method?

Example 2:
Document: bringing together the findings from interviews with stakeholders and research into the impact bond space conducted by the authors over the course of a year. In addition, the report draws on discussions from an intensive daylong workshop held in London in November 2016, in which impact bond practitioners from developing countries shared their experiences and early lessons learned. The report includes a Deal Book with detailed fact sheets for all impact bonds in developing countries, featuring both the four contracted and 24 in design phases, as of August 1, 2017.
Relevant Query: What is the research method?

Example 3:
Document: For each case, we conducted a document review of publicly available contract and loan agreements,1 press releases, and journalistic articles. We also conducted 11 semistructured interviews in January and February 2017,2 including five for the South Carolina SIB, three for the Chicago SIB, and three for the Utah SIB.
Relevant Query: What is the research method?

Example 4:
Document: Interviewees were selected based on their involvement in the design, launch, and implementation of the SIB. Interview questions covered program design, intervention scope, and management. During coding of these interviews, themes that emerged inductively shaped the four conceptual foci that guide our analysis: systemic change, performance metrics, cost structure, and social equity.
Relevant Query: What is the research method?

Example 5:
Document: The evidence base for this research wave is derived from the consultations and programme document review undertaken at the individual DIB level, the programme level and sector level.
Relevant Query: What is the research method?

Example 6:
Document: Internal learning workshops: The internal workshop brought together key stakeholders from across the three DFID DIB pilots and the Cameroon Cataract Bond. The workshop involved a discussion on the validity of these findings for the different DIBs, and additional perspectives and nuances across the range of DIBs present. Results from the learning workshop were used to refine the evaluation team's analysis and findings, and have been incorporated in this evaluation report.
Relevant Query: What is the research method?

Example 7:
Document: family- and center-based child care programs.
Relevant Query: What is the target population?

Example 8:
Document: approximately 18,500 families and 24,600 children
Relevant Query: What is the target population?

Example 9:

Document: Indigenous Asháninka people of the Ene River in the Peruvian Amazon, specifically members of the Kemito Ene producers association.
Relevant Query: What is the target population?

Example 10:
Document: The SIB will target unemployed vulnerable individuals who meet the following criteria used by Prosperidad Social (the sponsoring government entity): • Have a SISBEN score (poverty measure) of 0 to 41.74, are registered in Red Unidos (ultra-vulnerable group) or are victims of displacement due to the armed conflict; • Are between 18 and 40 years old; • Are high-school graduates; • Have not participated in Prosperidad Social's employment programs in the last two years.
Relevant Query: What is the target population?

Example 11:
Document: Three interviewees from confirming higher quality centers reported that accreditation was a challenge to advancing beyond a rating of 3, with two directors expressly men- tioning the challenge in paying for accreditation.
Relevant Query: What are the costs of the contract?

Example 12:
Document: OUTCOME FUNDS (USD) $110,000
Relevant Query: What are the costs of the contract?

Example 13:
Document: PAYMENT SCHEDULE AND AMOUNTS: Outcome funder pays investor at end of pilot for each outcome metric: $27,500 payment if 100% of target achieved, $20,625 payment if 75% of target achieved, $13,750 payment if 50% of target achieved, no payment if target not achieved.
Relevant Query: What are the costs of the contract?

Example 14:
Document: Results: On average, students in EG schools gained an additional 1.08 ASER levels compared to students in control schools ($p < 0.01$). Differences in aggregate learning gains between treatment and control schools were much greater in Year 3 (+6,045 learning levels) than in Year 2 (+1,434 levels) or in Year 1 (+1,461 levels).3
Relevant Query: What impact was achieved?

Example 15:
Document: Learning Gains against the DIB Target Students in EG schools gained on average an additional 1.08 ASER learning levels compared to students in control schools ($p < 0.01$).14 Learning gains for students in EG schools are 28% or 0.31 standard deviations larger than gains for students in control schools, comparing favorably with primary school programs aimed at improving test scores in rural India.15
Relevant Query: What impact was achieved?

Example 16:
Document: By the end of the three-year project, Educate Girls had enrolled 768 out-of-school girls, representing 92% of all identified out-of-school school girls eligible for enrollment. Educate Girls thus exceeded the enrollment target of 79% by 16%.
Relevant Query: What impact was achieved?

Example 17:
Document: {document_text}
Relevant Query:

**Appendix C: IR Model Results**

| | *Study Design* Results (Precision@20 / Recall @20) | | | |
|---|---|---|---|---|
| Paper | What is the study design? | What is the research method? | How was data collected and analysed? | Keywords (Concatenation) |
| **Model 1 Performance** | | | | |
| #2598 | 0.35/0.83 | 0.35/0.83 | 0.40/0.83 | 0.45/0.83 |
| #17247 | 0.00/0.00 | 0.00/0.00 | 0.15/1.00 | 0.15/1.00 |
| #17284 | 0.30/1.00 | 0.30/1.00 | 0.30/1.00 | 0.30/1.00 |
| #17755 | 0.00/0.00 | 0.25/0.33 | 0.00/0.00 | 0.20/0.33 |
| #17192 | 0.00/0.00 | 0.00/0.00 | 0.00/0.00 | 0.00/0.00 |
| #17725 | 0.00/0.00 | 0.05/1.00 | 0.05/1.00 | 0.05/1.00 |
| **Model 2 Performance** | | | | |
| #2598 | 0.20/0.33 | 0.30/0.67 | 0.15/0.33 | 0.20/0.33 |
| #17247 | 0.00/0.00 | 0.00/0.00 | 0.00/0.00 | 0.00/0.00 |
| #17284 | 0.00/0.00 | 0.00/0.00 | 0.00/0.00 | 0.30/1.00 |
| #17755 | 0.05/0.11 | 0.00/0.00 | 0.20/0.22 | 0.10/0.33 |
| #17192 | 0.00/0.00 | 0.00/0.00 | 0.00/0.00 | 0.00/0.00 |
| #17725 | 0.05/1.00 | 0.05/1.00 | 0.05/1.00 | 0.05/1.00 |
| **Model 3 Performance** | | | | |
| #2598 | 0.45/1.00 | 0.50/1.00 | 0.30/0.75 | 0.30/0.62 |
| #17247 | 0.15/1.00 | 0.15/1.00 | 0.15/1.00 | 0.00/0.00 |
| #17284 | 0.30/1.00 | 0.30/1.00 | 0.30/1.00 | 0.30/1.00 |
| #17755 | 0.05/0.05 | 0.10/0.09 | 0.30/0.36 | 0.15/0.14 |
| #17192 | 0.25/0.56 | 0.00/0.00 | 0.20/0.44 | 0.20/0.44 |
| #17725 | 0.05/1.00 | 0.05/1.00 | 0.05/1.00 | 0.05/1.00 |
| **Model 4 Performance** | | | | |
| #2598 | 0.50/1.00 | 0.40/0.88 | 0.40/0.88 | 0.25/0.50 |
| #17247 | 0.00/0.00 | 0.00/0.00 | 0.15/1.00 | 0.00/0.00 |
| #17284 | 0.30/1.00 | 0.30/1.00 | 0.30/1.00 | 0.30/1.00 |
| #17755 | 0.00/0.00 | 0.20/0.18 | 0.30/0.23 | 0.10/0.18 |
| **#17192** | 0.15/0.33 | 0.15/0.33 | 0.05/0.11 | 0.05/0.11 |
| **#17725** | 0.05/1.00 | 0.05/1.00 | 0.05/1.00 | 0.05/1.00 |

**Table C.1**: IR Results for 'Study Design' theme across the 4 models using sentences as retrieval unit

| Target Population Results (Precision@20 / Recall @20) | | | | | |
|---|---|---|---|---|---|
| Paper | What is the target population? | Who are the intended beneficiaries of the service? | Who does the service try to help? | Who was eligible for inclusion in the intervention? | Keywords (Concatenation) |
| **Model 1 Performance** | | | | | |
| **#2598** | 0.05/0.33 | 0.05/0.33 | 0.00/0.00 | 0.05/0.33 | 0.05/0.33 |
| **#17247** | 0.15/0.04 | 0.15/0.04 | 0.00/0.00 | 0.00/0.00 | 0.15/0.04 |
| **#17284** | 0.20/1.00 | 0.05/0.50 | 0.00/0.00 | 0.05/0.50 | 0.05/0.50 |
| **#17755** | 0.05/0.14 | 0.00/0.00 | 0.00/0.00 | 0.00/0.00 | 0.00/0.00 |
| **#17192** | 0.05/0.50 | 0.05/0.50 | 0.00/0.00 | 0.15/1.00 | 0.10/1.00 |
| **#17725** | N/A | N/A | N/A | N/A | N/A |
| **Model 2 Performance** | | | | | |
| **#2598** | 0.05/0.33 | 0.05/0.33 | 0.00/0.00 | 0.05/0.33 | 0.05/0.33 |
| **#17247** | 0.15/0.04 | 0.00/0.00 | 0.00/0.00 | 0.00/0.00 | 0.15/0.04 |
| **#17284** | 0.00/0.00 | 0.00/0.00 | 0.00/0.00 | 0.00/0.00 | 0.00/0.00 |
| **#17755** | 0.00/0.00 | 0.00/0.00 | 0.00/0.00 | 0.00/0.00 | 0.00/0.00 |
| **#17192** | 0.00/0.00 | 0.00/0.00 | 0.00/0.00 | 0.10/1.00 | 0.05/0.50 |
| **#17725** | N/A | N/A | N/A | N/A | N/A |
| **Model 3 Performance** | | | | | |
| **#2598** | .15/1.00 | 0.15/1.00 | 0.10/0.67 | 0.15/1.00 | 0.10/0.67 |
| **#17247** | 0.15/0.03 | 0.20/0.03 | 0.15/0.03 | 0.25/0.03 | 0.15/0.03 |
| **#17284** | 0.25/1.00 | 0.20/1.00 | 0.25/1.00 | 0.20/1.00 | 0.25/1.00 |
| **#17755** | 0.10/0.57 | 0.10/0.43 | 0.10/0.43 | 0.00/0.00 | 0.05/0.29 |
| **#17192** | 0.05/0.33 | 0.05/0.33 | 0.05/0.33 | 0.15/1.00 | 0.05/0.33 |
| **#17725** | N/A | N/A | N/A | N/A | N/A |
| **Model 4 Performance** | | | | | |
| **#2598** | 0.10/0.67 | 0.05/0.33 | 0.05/0.33 | 0.15/0.67 | 0.05/0.33 |
| **#17247** | 0.15/0.03 | 0.15/0.03 | 0.15/0.03 | 0.15/0.03 | 0.15/0.03 |
| **#17284** | 0.20/1.00 | 0.20/1.00 | 0.15/0.50 | 0.15/0.50 | 0.20/1.00 |
| **#17755** | 0.10/0.57 | 0.10/0.43 | 0.10/0.43 | 0.00/0.00 | 0.10/0.43 |
| **#17192** | 0.05/0.33 | 0.00/0.00 | 0.00/0.00 | 0.15/1.00 | 0.00/0.00 |
| **#17725** | N/A | N/A | N/A | N/A | N/A |

**Table C.2**: IR Results for 'Target Population' theme across the 4 models using sentences as retrieval unit

| | *Financial Detail and Costs* Results (Precision@20 / Recall @20) | | | | | |
|---|---|---|---|---|---|---|
| Paper | What are the costs of the contract? | How much is paid for outcomes? | What are the outcomes payments? | What is the total contract value? | What is the price per outcome? | Keywords (Concatenation) |
| **Model 1 Performance** | | | | | | |
| #2598 | 0.10/0.50 | 0.00/0.00 | 0.00/0.00 | 0.00/0.00 | 0.00/0.00 | 0.00/0.00 |
| #17247 | 0.00/0.00 | 0.15/0.10 | 0.15/0.10 | 0.05/0.10 | 0.00/0.00 | 0.05/0.10 |
| #17284 | 0.10/0.50 | 0.10/0.50 | 0.10/0.50 | 0.10/0.50 | 0.10/0.50 | 0.15/1.00 |
| #17755 | 0.75/0.40 | 0.45/0.40 | 0.15/0.20 | 0.60/0.33 | 0.45/0.40 | 0.40/0.40 |
| #17192 | N/A | N/A | N/A | N/A | N/A | N/A |
| #17725 | N/A | N/A | N/A | N/A | N/A | N/A |
| **Model 2 Performance** | | | | | | |
| #2598 | 0.10/0.50 | 0.00/0.00 | 0.00/0.00 | 0.10/0.50 | 0.00/0.00 | 0.00/0.00 |
| #17247 | 0.00/0.00 | 0.00/0.00 | 0.00/0.00 | 0.00/0.00 | 0.00/0.00 | 0.05/0.10 |
| #17284 | 0.10/0.50 | 0.10/0.50 | 0.10/0.50 | 0.10/0.50 | 0.00/0.00 | 0.10/0.50 |
| #17755 | 0.70/0.27 | 0.65/0.47 | 0.10/0.20 | 0.15/0.07 | 0.05/0.20 | 0.35/0.13 |
| #17192 | N/A | N/A | N/A | N/A | N/A | N/A |
| #17725 | N/A | N/A | N/A | N/A | N/A | N/A |
| **Model 3 Performance** | | | | | | |
| #2598 | 0.05/0.25 | 0.00/0.00 | 0.00/0.00 | 0.00/0.00 | 0.00/0.00 | 0.00/0.00 |
| #17247 | 0.00/0.00 | 0.30/0.15 | 0.15/0.12 | 0.10/0.15 | 0.10/0.12 | 0.05/0.12 |
| #17284 | 0.10/0.67 | 0.00/0.00 | 0.00/0.00 | 0.10/0.67 | 0.00/0.00 | 0.00/0.00 |
| #17755 | 0.05/0.03 | 0.30/0.22 | 0.15/0.09 | 0.10/0.06 | 0.00/0.00 | 0.00/0.00 |
| #17192 | N/A | N/A | N/A | N/A | N/A | N/A |
| #17725 | N/A | N/A | N/A | N/A | N/A | N/A |
| **Model 4 Performance** | | | | | | |
| #2598 | 0.30/1.00 | 0.10/0.25 | 0.10/0.25 | 0.10/0.25 | 0.25/1.00 | 0.00/0.00 |
| #17247 | 0.05/0.03 | 0.25/0.15 | 0.10/0.12 | 0.15/0.15 | 0.00/0.00 | 0.00/0.00 |
| #17284 | 0.10/0.67 | 0.10/0.67 | 0.10/0.67 | 0.10/0.67 | 0.00/0.00 | 0.10/0.67 |
| #17755 | 0.60/0.22 | 0.15/0.09 | 0.15/0.09 | 0.40/0.25 | 0.05/0.03 | 0.00/0.00 |
| #17192 | N/A | N/A | N/A | N/A | N/A | N/A |
| #17725 | N/A | N/A | N/A | N/A | N/A | N/A |

**Table C.3**: IR Results for 'Financial Detail and Costs' theme across the 4 models using sentences as retrieval unit

| | *Personal-level outcomes* Results (Precision@20 / Recall @20) | | | | | |
|---|---|---|---|---|---|---|
| Paper | What outcomes were achieved? | What impact was achieved? | What were the results of the intervention? | What was the impact of the intervention? | Were the contracted outcomes achieved? | Keywords (Concatenation) |
| **Model 1 Performance** | | | | | | |
| #2598 | N/A | N/A | N/A | N/A | N/A | N/A |
| #17247 | 0.00/0.00 | 0.00/0.00 | 0.00/0.00 | 0.00/0.00 | 0.00/0.00 | 0.00/0.00 |
| #17284 | 0.05/1.00 | 0.05/1.00 | 0.05/1.00 | 0.05/1.00 | 0.00/0.00 | 0.05/1.00 |
| #17755 | N/A | N/A | N/A | N/A | N/A | N/A |
| #17192 | 0.00/0.00 | 0.00/0.00 | 0.10/0.33 | 0.00/0.00 | 0.00/0.00 | 0.00/0.00 |
| #17725 | 0.40/0.56 | 0.45/0.78 | 0.40/0.56 | 0.35/0.56 | 0.65/1.00 | 0.45/0.78 |
| **Model 2 Performance** | | | | | | |
| #2598 | N/A | N/A | N/A | N/A | N/A | N/A |
| #17247 | 0.00/0.00 | 0.00/0.00 | 0.00/0.00 | 0.00/0.00 | 0.00/0.00 | 0.00/0.00 |
| #17284 | 0.00/0.00 | 0.05/1.00 | 0.00/0.00 | 0.05/1.00 | 0.05/1.00 | 0.00/0.00 |
| #17755 | N/A | N/A | N/A | N/A | N/A | N/A |
| #17192 | 0.00/0.00 | 0.00/0.00 | 0.10/0.33 | 0.00/0.00 | 0.00/0.00 | 0.00/0.00 |
| #17725 | 0.35/0.56 | 0.10/0.11 | 0.30/0.56 | 0.10/0.11 | 0.25/0.44 | 0.35/0.56 |
| **Model 3 Performance** | | | | | | |
| #2598 | N/A | N/A | N/A | N/A | N/A | N/A |
| #17247 | 0.00/0.00 | 0.00/0.00 | 0.00/0.00 | 0.00/0.00 | 0.00/0.00 | 0.00/0.00 |
| #17284 | 0.05/1.00 | 0.05/1.00 | 0.00/0.00 | 0.05/1.00 | 0.05/1.00 | 0.05/1.00 |
| #17755 | N/A | N/A | N/A | N/A | N/A | N/A |
| #17192 | 0.20/0.44 | 0.35/0.78 | 0.00/0.00 | 0.00/0.00 | 0.00/0.00 | 0.15/0.33 |
| #17725 | 0.45/0.79 | 0.50/0.79 | 0.10/0.14 | 0.10/0.14 | 0.10/0.21 | 0.45/0.79 |
| **Model 4 Performance** | | | | | | |
| #2598 | N/A | N/A | N/A | N/A | N/A | N/A |
| #17247 | 0.00/0.00 | 0.00/0.00 | 0.00/0.00 | 0.00/0.00 | 0.00/0.00 | 0.00/0.00 |
| #17284 | 0.00/0.00 | 0.05/1.00 | 0.00/0.00 | 0.05/1.00 | 0.00/0.00 | 0.05/1.00 |
| #17755 | N/A | N/A | N/A | N/A | N/A | N/A |
| #17192 | 0.10/0.22 | 0.00/0.00 | 0.10/0.22 | 0.00/0.00 | 0.00/0.00 | 0.00/0.00 |
| #17725 | 0.35/0.57 | 0.30/0.57 | 0.35/0.57 | 0.35/0.57 | 0.60/1.00 | 0.55/1.00 |

**Table C.4**: IR Results for 'Personal-level Outcomes' theme across the 4 models using sentences as retrieval unit

**Appendix D: Interviews participants**

**Theme: Background**
**Question: Interviewee's background and your current role?**

**i1**: Assistant Professor and Researcher at a University in Brazil. (Note: he had a previous background on outcomes contracts)

**i2**: Data Scientist Oxford Covid Tracker at Blavatnik School of Government (BSG). Previously, Research Assistant at BSG. Currently doing a PhD in FGV - Brazil, São Paulo. (Note: he had NO previous background on outcomes contracts)

**i3**: Research and Policy Associate at the Government Outcomes Lab, Blavatnik School of Government (BSG). Interested in relational contracting and other forms of partnerships. (Note: he had a previous background on outcomes contracts)

**i4**: Data Analyst in 'Pay4Fun' (a gateway for payments). He used to work for INSPER Metricis (Brazil), as a Research Assistant. He worked with ml techniques to extract information on outcomes-based contracts from different online sources. He is studying computer engineering. (Note: he had a previous background on outcomes contracts)

**i5**: MPP student working with the australian government, innovative funding and financing structures; Interested in facilitating emerging technologies and exploring the potential of new technologies to enable and empower people. (Note: he had no previous background on outcomes contracts)

**i6**: Lecturer in Organisational Management. She was a previous research fellow at GO Lab and her research is focused on public sector contracting and partnerships. For instance, how does government work with other entities to carry out their work? and how do they manage their relationships?

**Theme: Use of evidence in day-to-day job**
**Question: What does evidence look like on your day to day job?**

**i1**: Academic papers and literature reviews. He looks for new academic papers in Google scholar. When he reads a paper, he usually looks at the references and read those papers too. He also works with raw data that could come from sources like the World Bank Indicators or the Brazilian Statistics Office.

**i2**: The Oxford Covid tracker is tracking government responses to the COVID pandemic. A set of volunteers check how different governments react to the pandemic and share that dat with the Oxford COVID tracker. The OXCT collates all that data together and then work to analyse the whole dataset and present some data viz. The pay special attention to pandemic responses and policies, so they look at newspapers, news outlets, regulations, laws, decrees, etc.

**i3**: Evidence in his day-to-day job: primary data, or existing academic literature. More practice-focused evaluations. Examples of primary sources: interviews, focus groups, observations. Secondary data: archives, academic articles, policy evaluation reports, etc. Source of evidence: Google scholar and other academic search engines. National Archives (UK government - online version).

**i4**: Data and evidence coming from a Google search, which has advantages and limitations. Sometimes no relevant data. He would also use tools like: https://mention.com/en/

**i5**: Desktop research, collecting documents from Australia and the UK, consulting external partners with expertise, observing the development of new tools, and obtaining more detailed information about the impacts and implications of new technologies

**i6**: Internal performance publications, publicly available literature, for example, information from the Urban Institute. I also use Google Scholar, I also set up notifications and alerts for new findings